\documentclass[11pt]{article}

\usepackage[preprint]{acl}

\usepackage{times}
\usepackage{latexsym}

\usepackage[T1]{fontenc}

\usepackage[utf8]{inputenc}

\usepackage{microtype}

\usepackage{inconsolata}

\usepackage{graphicx}

\title{The Truth Beneath the Tokens: Prompt-Space Recovery of Knowledge Edits in Language Models}

\author{First Author \\
  Affiliation / Address line 1 \\
  Affiliation / Address line 2 \\
  Affiliation / Address line 3 \\
  \texttt{email@domain} \\\And
  Second Author \\
  Affiliation / Address line 1 \\
  Affiliation / Address line 2 \\
  Affiliation / Address line 3 \\
  \texttt{email@domain} \\}

\usepackage{times,latexsym}
\usepackage{url}
\usepackage{xcolor}
\usepackage[most]{tcolorbox}
\usepackage{ragged2e}
\usepackage{amsfonts}
\usepackage{booktabs}
\usepackage{multirow}
\usepackage{tabularx}
\usepackage[table]{xcolor}

\usepackage{enumitem}
\usepackage{listings}

\usepackage[draft]{labreview}

\newtcolorbox{promptbox}{
    colback=gray!4,
    colframe=gray!45,
    boxrule=0.5pt,
    arc=2mm,
    left=2mm,
    right=2mm,
    top=2mm,
    bottom=2mm,
    breakable,
    fontupper=\small
}

\usepackage{xspace,mfirstuc,tabulary}

\newcommand{\gain}[1]{\textcolor{green!60!black}{\scriptsize{(+#1)}}}
\newcommand{\loss}[1]{\textcolor{red!70!black}{\scriptsize{(-#1)}}}

\definecolor{LlamaGreen}{RGB}{81,129,80}
\definecolor{GPTJBlue}{RGB}{16,110,169}
\definecolor{GPTRed}{RGB}{255,51,29}

\newcommand{\llama}{{\textcolor{LlamaGreen}{Llama3}}\xspace}
\newcommand{\gpt}{{\textcolor{GPTJBlue}{GPT2-XL}}\xspace}
\newcommand{\gptj}{{\textcolor{GPTRed}{GPT-J}}\xspace}

\newtcolorbox{qualemail}[3][]{
  enhanced,
  colback=black!1,
  colframe=#3!25!white,
  colbacktitle=#3!5!white,
  coltitle=black!70,
  title={#2},
  fonttitle=\mdseries\normalsize,
  boxrule=0.5pt,
  arc=1.5mm,
  left=1.7mm,
  right=1.7mm,
  top=1.2mm,
  bottom=1.2mm,
  before upper={\justifying\normalsize\setlength{\parindent}{0pt}\setlength{\parskip}{2pt}},
  #1
}

\newcommand{\Memento}{\textsc{Memento}\xspace}

\author{Manit Baser$^1$, Rachel Yeo Hui Min$^1$, Aditya Nawal$^1$, Dinil Mon Divakaran$^{1,2}$, Mohan Gurusamy$^1$ \\
  $^1$National University of Singapore, Singapore \\
  $^2$A*STAR Institute of Advanced Intelligence and Computing (A*STAR IAIC), Singapore}

\date{}

\begin{document}
\maketitle
\begin{abstract}

Model editing provides an efficient way to update factual knowledge in transformer models. Recent approaches reverse edits by modifying, masking, or decomposing model weights, but make recovery dependent on weight-space interventions. We instead show that pre-edit knowledge can be recovered through input-space interventions alone, without modifying the edited model's weights. We introduce \textbf{\Memento}, a prompt-space mechanism for auditing residual knowledge that learns a reusable recovery signal to steer an edited model toward its pre-edit response. \Memento supports both continuous recovery prefixes and discrete textual prompts for standard text interfaces. Across three LLMs and five editing techniques, \Memento recovers 80.07\% of held-out edited facts, achieving a 9.58\% relative improvement over baselines, suggesting that editing may suppress the facts without fully eliminating their recoverability. We further investigate whether prompt-space recovery extends beyond model editing. On pristine, unedited LLMs, \Memento improves average factual recall by 53.9\% relative to standard inference. These results demonstrate the potential of prompt-space recovery for auditing residual knowledge in edited models and investigating the gap between factual knowledge and its accessibility during inference. We will release our source code.
\end{abstract}

\section{Introduction}

Model editing is used for correcting factual content in large language models (LLMs) after training, including removal of personally identifiable information~\cite{ruzzetti-etal-2025-private, venditti2026enhancing}, and correction of outdated or incorrect facts~\cite{meng2022locating, meng2023massediting}. Retraining from scratch offers the strongest guarantee but is infeasible at deployment scale~\cite{hu2024duty}; fine-tuning is more practical but remains expensive for large models~\cite{fang2025alphaedit}; and retrieval-augmented generation moves knowledge control outside the model, introducing failure modes of its own~\cite{an2025rag, cho2024typos}. Model editing instead performs precise, localized parameter updates, typically following a locate-then-edit paradigm, where it identifies the weights responsible for a target association, and then rewrites them.

The success of an edit is typically evaluated behaviorally, that is, if the model produces the new target when prompted, the edit is considered successful~\cite{meng2022locating, zhong2023mquake}. This has also motivated a line of work on edit reversal. One Mask~\cite{holmov-etal-2026-one} learns a shared sparse binary mask over edited MLP weights to reverse edits across multiple facts. \citet{youssef2026tracing} show that edited information is concentrated in the dominant singular subspace of the edited weight matrix. \citet{jiang2026selective} study selective reversal by learning element-wise gates over singular vectors to reverse targeted facts while preserving other edits. These approaches, however, formulate recovery as weight-space reversal, modifying model weights to undo the effects of an edit. Consequently, they do not establish whether pre-edit knowledge remains recoverable from the edited model without reversing the parameter update. This leaves a fundamental question unexplored: does editing truly eliminate the pre-edit knowledge, or does it primarily suppress its accessibility under standard prompting?

We introduce \textbf{\Memento}, a novel prompt-space auditing mechanism for recovering pre-edit knowledge from edited LLMs. Rather than modifying model parameters, \Memento learns a recovery signal that steers an edited model toward its pre-edit factual response. \Memento can be trained in both continuous and discrete prompt spaces. To elicit the pre-edit response, the continuous variant learns a soft input representation, while the discrete variant learns a textual recovery prompt that is prepended to the input.

Across three models and five editing techniques, continuous \Memento recovers 80.07\% of held-out edited facts, achieving a 9.58\% relative improvement over One Mask~\cite{holmov-etal-2026-one}. Our discrete variant achieves 49.0\% recovery using a textual recovery prompt, making it compatible with standard API interfaces once the prompt has been learned. Together, these results suggest that editing may suppress factual knowledge without fully eliminating its recoverability.


Finally, we investigate whether \Memento can surface factual knowledge beyond editing. Recent work~\cite{calderon2026empty} distinguishes factual knowledge encoded in model weights from what is accessible through standard inference. Building on this distinction, we evaluate whether a learned prompt-space recovery signal can improve factual recall in pristine, unedited models. Across three LLMs, \Memento improves average recall by 53.9\% relative to standard inference. While prior work characterizes the gap between factual encoding and recall, our results show that a reusable prompt-space intervention elicits facts that standard prompting fails to retrieve.

Our contributions are summarized as follows:
\begin{enumerate}
    \item
    We introduce \Memento, a prompt-space auditing mechanism to recover pre-edit knowledge in LLMs without modifying their parameters. \Memento learns reusable recovery signals in both continuous and discrete prompt spaces, including a textual recovery prompt that can be supplied through standard language-model interfaces.

    \item
    We evaluate \Memento on multiple LLMs and editing techniques. The continuous variant recovers 80.07\% of the held-out facts, a 9.58\% relative improvement over the baseline, while the discrete variant achieves 49.0\% recovery, showing that pre-edit knowledge can be recovered even via textual signal.

    \item
    We show that \Memento can also improve factual recall in pristine, unedited LLMs. Across three models, it improves average recall by 53.9\% relative to standard inference, suggesting that prompt-space recovery can help probe the gap between factual knowledge a model may contain and what it retrieves under standard prompting.
\end{enumerate}




\begin{figure*}
    \centering
    \includegraphics[width=0.99\linewidth]{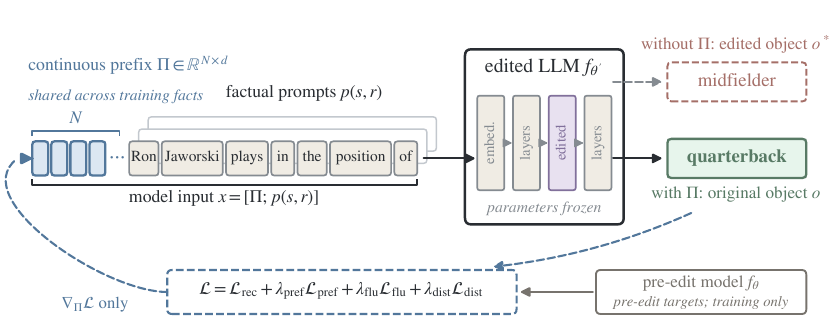}
    \caption{Overview of \Memento's continuous prompt-space recovery. A trainable prefix $\Pi$ is prepended to factual prompts and optimized across training facts to recover the original object $o$ from an edited model $f_{\theta'}$, whose parameters remain frozen. The objective combines supervised recovery ($\mathcal{L}_{\mathrm{rec}}$), preference optimization ($\mathcal{L}_{\mathrm{pref}}$), fluency preservation ($\mathcal{L}_{\mathrm{flu}}$), and distillation from the pre-edit model ($\mathcal{L}_{\mathrm{dist}}$). Gradients update only $\Pi$; the pre-edit model supplies distillation targets during training and is not needed at inference. Without the prefix, the edited LLM answers the edited object ($o^*$ = \textit{midfielder}); with it, the LLM produces the original object ($o$ = \textit{quarterback}).}
\label{fig:memento}
\end{figure*}

\section{Related Works}
\label{sec:related_work}

\paragraph{Model Editing.}
Model editing aims to update factual associations in LLMs without retraining from scratch. Editing techniques divide into parameter-preserving approaches, which add memory modules or rely on in-context abilities~\cite{mitchell2022memory,hartvigsen2023aging,zheng2023can}, and parameter-modifying techniques, like ROME and MEMIT~\cite{meng2022locating, meng2023massediting}, which modify the weights. Subsequent approaches improve edit locality and preservation through null-space projection, weight regularization, and constrained matrix updates~\cite{fang2025alphaedit,gu2024rect,ma2024perturbation}. However, works on ripple effects and multi-hop reasoning further show editing can have unintended consequences beyond targeted facts~\cite{cohen2024evaluating,baser-etal-2026-clare,baser2026thinkeval}. We study the parameter-modifying family, since an edit that never enters the weights raises no question about what the weights retain.

\paragraph{Edit Reversal and Recovery.}
Recent work investigates if model edits can be reversed to restore pre-edit behavior. One Mask~\cite{holmov-etal-2026-one} learns a shared sparse mask over edited weights to reverse edits across multiple facts. \citet{youssef2026tracing} propose a reversal method based on bottom-rank approximations of edited weight matrices. \citet{jiang2026selective} introduce gated singular-vector shrinkage for selectively reversing edits. These methods suggest that a lot of pre-edit behavior can be recovered via weight-space interventions. Prompt-based reversal has been explored for in-context edits~\cite{youssef-etal-2025-make}. \citet{baser2026tokens} recover pre-edit responses by searching over alternative tokenizations of a query without modifying the weights. However, its reconstruction procedure incurs additional inference overhead, requiring multiple generations over alternative tokenizations for each target fact.


\paragraph{Prompt-Based Optimization.}
Prompt-based optimization steers frozen LLMs through optimized input representations. Continuous methods, such as prompt tuning and prefix tuning, learn trainable input embeddings while keeping the weights fixed~\cite{li-liang-2021-prefix,lester-etal-2021-power}. Discrete methods optimize sequences of vocabulary tokens. For example, \citet{zou2023universal} use gradient-guided search to construct adversarial suffixes that elicit targeted behaviors. \citet{yu-etal-2026-need} further show that optimizing a single continuous prefix embedding can refine model behavior while keeping the backbone frozen. \Memento adopts prompt-based optimization for targeting recovery of pre-edit knowledge. It learns a continuous recovery prefix or searches for a discrete textual signal, evaluating whether the signal generalizes to different editing techniques.

\section{\Memento: Prompt-Space Recovery of Edited Knowledge}
\label{sec:method}

\Memento is a prompt-space approach to recovering pre-edit factual behavior from LLMs without modifying their parameters. As illustrated in Figure~\ref{fig:memento}, it learns a reusable recovery signal in continuous or discrete prompt space, optimized to restore original factual associations.

We represent a factual association as a triple $(s,r,o)$, where $s$ denotes a subject, $r$ a relation, and $o$ the original object, and $p(s,r)$ denotes the corresponding factual prompt. Given a model $f_{\theta}$ and an editing request $e=(s,r,o,o^*)$, where $o^*$ denotes the target object that replaces $o$, a model editing technique $\mathcal{E}$ produces an edited model
\begin{equation}
    f_{\theta'} = \mathcal{E}(f_{\theta}, e),
\end{equation}
such that the edited model preferentially predicts $o^*$ in response to $p(s,r)$. Our goal is to recover the original object $o$ from the edited model without modifying its weights. We refer to this as \textit{knowledge recovery}, distinguishing it from weight-space edit reversal, which intervenes directly on the model weights to counteract the effects of an edit. We analyze training and per-query costs, together with their asymptotic complexity, in Appendix~\ref{app:cost}.

\subsection{Prompt-Space Recovery}
\label{sec:continuous}

We hypothesize that editing can suppress the retrieval of a factual association without completely eliminating its accessibility. If so, an appropriate input intervention may be able to elicit the original association. \Memento learns a continuous recovery prefix $\Pi \in \mathbb{R}^{N \times d}$, where $N$ is the prefix length and $d$ is the model's embedding dimension. Given the input embedding matrix $W_E$ and tokenized factual prompt $q$, the resulting input is
\begin{equation}
    x = [\Pi; W_E[q]].
\end{equation}

where $[\,;\,]$ denotes concatenation. A single prefix shared across a collection of edited facts $D_{\mathrm{train}}$ and optimized with an objective function $\mathcal{L}$ as

\begin{equation}
\begin{aligned}
\Pi^*
&= \arg\min_{\Pi}\;
\mathbb{E}_{(s,r,o,o^*)\sim D_{\mathrm{train}}}
\Big[
\\[-2pt]
&\qquad\mathcal{L}(\Pi;s,r,o,o^*)
\Big],
\end{aligned}
\label{eq:universal}
\end{equation}
This prefix captures a shared recovery direction rather than storing the answer to individual edits.

\subsection{Continuous Recovery Training Objective}
\label{sec:objective}

We train the continuous recovery prefix using four complementary loss terms, each serving a distinct purpose:
(i) Recovery loss ($\mathcal{L}_{\mathrm{rec}}$) increases the likelihood of the pre-edit response;
(ii) Preference loss ($\mathcal{L}_{\mathrm{pref}}$) explicitly encourages the pre-edit response to be preferred over the edited target;
(iii) Fluency loss ($\mathcal{L}_{\mathrm{flu}}$) limits changes to the model's behavior on inputs unrelated to the edited facts; and
(iv) Distillation loss ($\mathcal{L}_{\mathrm{dist}}$) aligns the edited model's predictions under the recovery prefix with those of the pre-edit model.
Together, these objectives encourage factual recovery while constraining unintended changes to model behavior.

For a prefix $z$ and a continuation $y$, we define the length-normalized log-likelihood under $f_{\theta'}$ as
\begin{equation}
    \ell(y \mid z)
    =
    \frac{1}{|y|}
    \sum_i
    \log P_{\theta'}
    \left(
        y_i \mid z,p(s,r),y_{<i}
    \right),
\end{equation}
where $z=\Pi$ for continuous recovery. Let $\mathbf{o}$ and $\mathbf{o}^*$ denote the token sequences obtained by tokenizing the original object $o$ and edited target $o^*$, respectively. $\mathcal{L}_{\mathrm{rec}}$ increases the likelihood of $o$:
\begin{equation}
    \mathcal{L}_{\mathrm{rec}}
    =
    -\ell(\mathbf{o}\mid z).
\end{equation}
However, increasing its likelihood alone does not explicitly encourage the original object to be preferred over the edited target. Hence, we introduce $\mathcal{L}_{\mathrm{pref}}$ to separate their likelihoods:
\begin{equation}
    \mathcal{L}_{\mathrm{pref}}
    =
    -\log \sigma
    \left(
        \beta\,\ell(\mathbf{o}\mid z)
        -
        \beta\,\ell(\mathbf{o}^*\mid z)
        -
        \gamma
    \right),
\end{equation}
where $\sigma(\cdot)$ denotes the sigmoid function, $\beta$ controls the preference scale, and $\gamma$ specifies the desired margin. $\mathcal{L}_{\mathrm{pref}}$ explicitly encourages the original object to receive a higher likelihood than the edited object, reducing the chance of the recovery prefix favoring the post-edit association.

We further introduce a fluency regularizer:
\begin{equation}
    \mathcal{L}_{\mathrm{flu}}
    =
    \mathbb{E}_{u}
    \left[
        D_{\mathrm{KL}}
        \left(
            P_{\theta'}(\cdot\mid u)
            \,\Vert\,
            P_{\theta'}(\cdot\mid \Pi,u)
        \right)
    \right],
\end{equation}
where $u$ denotes generic text unrelated to the edited facts, and $D_{\mathrm{KL}}(\cdot \Vert \cdot)$ denotes the Kullback-Leibler divergence from one probability distribution to another. $\mathcal{L}_{\mathrm{flu}}$ penalizes changes in output distribution caused by adding the recovery prefix to unrelated inputs. It thus discourages the prefix from altering behavior beyond the factual prompts for which recovery is intended.

We use $\mathcal{L}_{\mathrm{dist}}$ to align the edited model under the recovery prefix with the pre-edit model:
\begin{equation}
\begin{aligned}
\mathcal{L}_{\mathrm{dist}}
&= T^2 D_{\mathrm{KL}}\Big(
P_{\theta}^{T}(\cdot \mid p(s,r))
\,\Vert\, \\
&\qquad\qquad
P_{\theta'}^{T}(\cdot \mid \Pi,p(s,r))
\Big),
\end{aligned}
\end{equation}
where $P_{\theta}^{T}(\cdot\mid p(s,r))$ denotes the next-token probability distribution produced by the pre-edit model, with its logits scaled by the temperature $T$. $P_{\theta'}^{T}(\cdot\mid \Pi,p(s,r))$ is defined analogously for the edited model with the recovery prefix. The KL divergence is computed over the vocabulary at the final prompt position, and $T$ controls the softness of the distributions. The overall objective is
\begin{equation}
    \mathcal{L}
    =
    \mathcal{L}_{\mathrm{rec}}
    + \lambda_{\mathrm{pref}}\mathcal{L}_{\mathrm{pref}}
    + \lambda_{\mathrm{flu}}\mathcal{L}_{\mathrm{flu}}
    + \lambda_{\mathrm{dist}}\mathcal{L}_{\mathrm{dist}}.
    \label{eq:objective}
\end{equation}
The coefficients $\lambda_{\mathrm{pref}}$, $\lambda_{\mathrm{flu}}$, and $\lambda_{\mathrm{dist}}$ control the relative contributions of the losses. We present their individual effects in Appendix~\ref{app:loss_ablation}.

\subsection{Discrete Recovery Training Objective}
\label{sec:discrete}

Let $\mathcal{V}$ be the model's token vocabulary. The continuous prefix is represented in the model's embedding space. To obtain a textual recovery prompt, \Memento learns a token sequence
\begin{equation}
    a=(a_1,\ldots,a_n),
    \qquad
    a_i\in\mathcal{V},
\end{equation}
which is prepended to the query. The prefix is optimized over the same collection of edited facts:
\begin{equation}
    \mathcal{L}_{\mathrm{disc}}(a)
    =
    \mathbb{E}_{(s,r,o,o^*)\sim D_{\mathrm{train}}}
    \left[
        \mathcal{L}_{\mathrm{rec}}
        +
        \lambda_{\mathrm{pref}}\mathcal{L}_{\mathrm{pref}}
    \right],
    \label{eq:discrete}
\end{equation}
where $z=a$ in likelihood terms. $\mathcal{L}_{\mathrm{disc}}(a)$ requires a single prefix to prefer original object over edited object across multiple facts. Let $\operatorname{dec}(\cdot)$ be the tokenizer's decoding function, and $\operatorname{tok}(\cdot)$ be the corresponding tokenization function. The search is constrained to text-realizable token sequences:

\begin{equation}
    \mathcal{A}
    =
    \left\{
        a\in\mathcal{V}^n:
        \operatorname{tok}(\operatorname{dec}(a))=a
    \right\},
\end{equation}
so that the learned sequence is preserved when decoded and re-tokenized by a text interface.

We optimize Eq.~\eqref{eq:discrete} over $\mathcal{A}$ using greedy coordinate gradient~\cite{zou2023universal}. The search iteratively proposes token substitutions ranked by the gradient with respect to one-hot token indicators, and evaluates feasible candidates exactly on batches of edited facts. The resulting sequence is the one with the lowest $\mathcal{L}_{\mathrm{disc}}$ on a subset of $D_{\mathrm{train}}$ withheld from the search. Hence, a single learned prefix can be used to elicit the pre-edit responses through a standard textual interface.

\section{Experiments}
\label{sec:experiments}

\paragraph{Training Details.}
We follow the setup of One Mask~\cite{holmov-etal-2026-one} and draw facts from the training and test splits of CounterFact~\cite{meng2022locating}. Following the configuration specified in the official One Mask repository, we use 1,000 facts for training and 1,000 held-out facts for evaluation. We evaluate five model editing techniques: MEMIT~\cite{meng2023massediting}, AlphaEdit~\cite{fang2025alphaedit}, PRUNE~\cite{ma2024perturbation}, RECT~\cite{gu2024rect}, and CoME~\cite{jung2025come}.

We evaluate \Memento on \gpt (1.5B), \gptj (6B), and \llama (8B, Instruct), and compare its recovery performance with One Mask. For the editing techniques, we first edit the model using the corresponding training set, and train the recovery prefix on the resulting edited model. The learned prefix is then evaluated on held-out facts from the test set (excluded from training).

We consider two evaluation settings. In the \emph{within-method} setting, the recovery prefix is trained and evaluated on models edited using the same editing technique. In the \emph{cross-method} setting, the prefix is trained using one technique and evaluated on a model edited using another. Specifically, we train recovery prefixes on each editing technique and evaluate them across other techniques. This tests whether the recovery signal captures properties shared across editing techniques rather than artifacts specific to a particular one. All hyperparameters and compute details for reproducibility are given in Appendix~\ref{app:reproducibility}.

\subsection{Evaluation Metrics}
\label{sec:metrics}

\textbf{Top-1 Recovery (Top-1).}
Our primary metric measures whether \Memento restores the pre-edit model’s top prediction, following One Mask~\cite{holmov-etal-2026-one}. For each factual prompt, we compare the top prediction of the edited model with the recovery prefix against that of the corresponding pre-edit model. We report the percentage of held-out facts for which the predictions match. Higher scores indicate more successful behavioral recovery. We use top-1 in both the within-method and cross-method settings, referring to the latter as \emph{Cross-Method Recovery (CMR)}.

\textbf{LLM-as-a-Judge Evaluation.}
For our research question 3, we evaluate factual recall on pristine, unedited models using an LLM judge. Given the factual prompt, its reference answer, and the model's generated continuation, the judge determines whether the continuation correctly expresses the requested fact. This semantic evaluation accommodates equivalent entity names and answers expressed indirectly, rather than requiring an exact lexical match. We use \texttt{deepseek-ai/DeepSeek-V4.1-Flash} as the judge and report the percentage of continuations judged correct. The model's generation uses FP32 inference and a maximum of 32 output tokens. Appendix~\ref{app:judge_evaluation} provides the judging instructions and illustrative examples.

\textbf{Canonical-Answer Correctness.}
For Research Question 3, we complement semantic judging with a stricter, rule-based metric. A continuation is counted as correct only if it begins with the canonical reference answer after Unicode, case, punctuation, and whitespace normalization. We report the percentage of continuations satisfying this criterion. This metric is deterministic and easy to reproduce, but may classify semantically valid answers as incorrect when they use aliases or express the answer indirectly.

\begin{table}
\centering
\small
\setlength{\tabcolsep}{6pt}
\caption{Efficacy of all editing techniques across the models being evaluated over the held-out facts.}
\label{tab:efficacy}
\begin{tabular}{l|ccc}
\toprule
\textbf{Technique}
& \gpt
& \gptj
& \llama\\
\midrule
AlphaEdit
& 0.97 & 0.97 & 0.92 \\
MEMIT
& 0.95 & 0.97 & 0.91 \\
RECT
& 0.90 & 0.95 & 0.85 \\
PRUNE
& 0.96 & 0.91 & 0.96 \\
CoME
& 0.95 & 0.94 & 0.90 \\
\bottomrule
\end{tabular}
\end{table}

\subsection{Editing Efficacy}
\label{sec:editing_efficacy}

Before evaluating recovery, we verify that the editing techniques successfully modify the models' factual predictions. Table~\ref{tab:efficacy} reports editing efficacy on the held-out facts across all three models and five editing techniques. All techniques achieve high efficacy, with scores ranging from 0.85 to 0.97. These results show that the evaluated models have been successfully edited, providing a meaningful basis for assessing whether \Memento can recover pre-edit behavior.

\subsection{Prefix Length Selection}
\label{subsec:length_selection}
The recovery prefix length $N$ determines the number of trainable input embeddings available to steer the edited model toward its pre-edit behavior. We select the prefix length using a calibration experiment with MEMIT-edited models. We evaluate $N \in \{1,2,4,8,16,32,64\}$ on all the three models, measuring top-1 recovery on each. As shown in Table~\ref{tab:prefix_calibration}, increasing the prefix length from $1$ to $2$ substantially improves recovery across all three models, indicating that a single input embedding provides limited capacity to recover the original factual associations. Performance generally improves with longer prefixes, although the gains are not monotonic and vary across models. In particular, $N=16$ achieves top-1 recovery scores of $0.79$, $0.80$, and $0.84$ on \gpt, \gptj, and \llama, respectively. While $N=8$ achieves a marginally higher score on \llama ($0.85$), $N=16$ provides consistently strong performance across all three architectures. Further increasing the prefix length to $32$ or $64$ does not yield consistent improvements, with the $64$-token prefix notably reducing recovery on \llama. Thus, we select $N=16$ as a common prefix length for all subsequent experiments, without introducing model-specific configurations.

\begin{table}
\centering
\small
\caption{Top-1 Recovery for different prefix lengths $N$ on MEMIT-edited models.}
\label{tab:prefix_calibration}
\begin{tabular}{c|ccc}
\toprule
$N$ & \gpt & \gptj & \llama \\
\midrule
1  & 0.58 & 0.54 & 0.57 \\
2  & 0.73 & 0.71 & 0.70 \\
4  & 0.71 & 0.68 & 0.77 \\
8  & 0.75 & 0.77 & \textbf{0.85} \\
16 & \textbf{0.79} & \textbf{0.80} & 0.84 \\
32 & 0.76 & 0.77 & 0.82 \\
64 & 0.77 & 0.72 & 0.58 \\
\bottomrule
\end{tabular}
\end{table}

\subsection{Recovery Effectiveness}
\label{sec:recovery}

\begin{qualemail}{Research Question 1}{black}
Can prompt-space interventions recover pre-edit behavior from knowledge-edited LLMs without modifying their weights?
\end{qualemail}

Table~\ref{tab:recovery_comparison} presents the results for within-method top-1 and cross-method recovery (CMR).

\paragraph{Within-Method Recovery.}
\Memento achieves strong recovery performance across the evaluated models, with particularly consistent improvements on \gptj and \llama, outperforming One Mask across all five editing techniques. The largest gain occurs with MEMIT on \llama, where recovery increases from 0.62 to 0.84. The results on \gpt are more mixed. \Memento improves top-1 recovery for AlphaEdit (0.78 to 0.80) and RECT (0.79 to 0.80), matches One Mask on MEMIT (0.79), and underperforms on PRUNE and CoME only by 5 and 6 percentage points respectively.

\begin{table*}[t]
\centering
\small
\setlength{\tabcolsep}{4pt}
\caption{Top-1 recovery and CMR of \Memento and One Mask across three LLMs and five editing techniques.}
\label{tab:recovery_comparison}
\begin{tabular}{ll|cc|cc|cc|cc|cc}
\toprule
\multirow{2.5}{*}{Model} & \multirow{2.5}{*}{Method}
& \multicolumn{2}{c|}{MEMIT}
& \multicolumn{2}{c|}{AlphaEdit}
& \multicolumn{2}{c|}{PRUNE}
& \multicolumn{2}{c|}{CoME}
& \multicolumn{2}{c}{RECT} \\
\cmidrule(lr){3-4} \cmidrule(lr){5-6}
\cmidrule(lr){7-8} \cmidrule(lr){9-10} \cmidrule(lr){11-12}
& & Top-1 & CMR
& Top-1 & CMR
& Top-1 & CMR
& Top-1 & CMR
& Top-1 & CMR \\
\midrule
\multirow{2}{*}{\gpt}
& One Mask
& \textbf{0.79} & 0.76
& 0.78 & 0.68
& \textbf{0.77} & \textbf{0.76}
& \textbf{0.83} & \textbf{0.78}
& 0.79 & \textbf{0.79} \\
& \Memento
& \textbf{0.79} & \textbf{0.77}
& \textbf{0.80} & \textbf{0.78}
& 0.72 & 0.74
& 0.77 & 0.75
& \textbf{0.80} & 0.78 \\
\midrule
\multirow{2}{*}{\gptj}
& One Mask
& 0.72 & 0.72
& 0.71 & 0.70
& 0.73 & 0.72
& 0.72 & 0.70
& 0.78 & 0.75 \\
& \Memento
& \textbf{0.80} & \textbf{0.78}
& \textbf{0.82} & \textbf{0.77}
& \textbf{0.74} & \textbf{0.77}
& \textbf{0.79} & \textbf{0.78}
& \textbf{0.81} & \textbf{0.82} \\
\midrule
\multirow{2}{*}{\llama}
& One Mask
& 0.62 & 0.56
& 0.78 & 0.68
& 0.64 & 0.57
& 0.65 & 0.59
& 0.65 & 0.69 \\
& \Memento
& \textbf{0.84} & \textbf{0.82}
& \textbf{0.85} & \textbf{0.83}
& \textbf{0.78} & \textbf{0.78}
& \textbf{0.84} & \textbf{0.80}
& \textbf{0.86} & \textbf{0.81} \\
\bottomrule
\end{tabular}
\end{table*}

\paragraph{Cross-Method Generalization.}
We examine whether recovery signals transfer across editing techniques. On \gptj and \llama, \Memento outperforms One Mask in CMR across all techniques, with gains ranging from 0.05 to 0.26. This suggests that the learned recovery signal can generalize beyond the editing procedure used to train it. On \gpt, the performance is again mixed. \Memento improves CMR for MEMIT and AlphaEdit, but performs slightly worse on the rest.

Overall, \Memento achieves competitive or better recovery than One Mask in most evaluated settings, with particularly strong results on \gptj and \llama. Crucially, \Memento achieves this recovery entirely through an input-space prefix, without modifying model weights. Thus, pre-edit behavior can be recovered through standard model inference, highlighting prompt-space recovery as a viable alternative for auditing model editing.

\begin{qualemail}{Key Takeaway}{teal}
\Memento achieves relative improvements of 9.58\% in top-1 recovery and 12.73\% in CMR over One Mask. This demonstrates pre-edit behavior can be recovered without modifying model weights, with recovery signals transferring across editing techniques.
\end{qualemail}

\subsection{Text-Only Recovery}
\label{sec:discrete_recovery}

\begin{qualemail}{Research Question 2}{black}
Can continuous recovery be expressed as a discrete prompt while retaining effectiveness?
\end{qualemail}

Unlike continuous prefixes, which operate directly in the embedding space, discrete recovery uses a sequence of ordinary text tokens that can be supplied through a standard language-model interface. Table~\ref{tab:discrete_recovery} reports the Top-1 recovery achieved by \Memento's discrete variant all the models and editing techniques. \gpt achieves the strongest results, reaching 0.56 with PRUNE.

These results show the practical effectiveness of discrete recovery for black-box auditing. By recovering pre-edit behavior through ordinary textual prompts, \Memento enables auditors to probe the persistence of edited knowledge using standard inference interfaces, without access to internal embeddings or weights. This makes recovery feasible in realistic, access-constrained settings and provides a measurable means of assessing whether model editing has successfully suppressed prior behavior or merely obscured it.

\begin{table*}
    \centering
    \caption{Factual recall on pristine, unedited models. \textit{Pristine Recall} column reports recall under standard inference. Remaining columns report recall using signals obtained under the indicated editing-technique settings. \texttt{deepseek-ai/DeepSeek-V4.1-Flash} is used as the LLM-as-a-Judge. Canonical-answer correctness requires canonical answer at the start of continuation after Unicode, case, punctuation, and whitespace normalization.}
    \label{tab:pristine_recall}
    \small
    \setlength{\tabcolsep}{4.5pt}
    \begin{tabular}{llcccccc}
        \toprule
        \multirow{2}{*}{\textbf{Model}} & \multirow{2}{*}{\textbf{Method}} & \textbf{Pristine} & \multicolumn{5}{c}{\textbf{Editing Technique Used to Obtain Recovery Signal}} \\
        \cmidrule(lr){4-8}
        & & \textbf{Recall} & \textbf{AlphaEdit} & \textbf{MEMIT} & \textbf{RECT} & \textbf{PRUNE} & \textbf{CoME} \\
        \midrule
        \multicolumn{8}{c}{\textbf{LLM-as-a-Judge Evaluation}} \\
        \midrule
        \multirow{2}{*}{\gpt}
        & One Mask & \multirow{2}{*}{\textbf{21.7\%}} & 12.5\% \loss{9.2}  & 12.2\% \loss{9.5}  & 11.4\% \loss{10.3} & 4.4\% \loss{17.3}  & 14.3\% \loss{7.4}  \\
        & \Memento &                                   & 33.3\% \gain{11.6} & 39.2\% \gain{17.5} & 47.0\% \gain{25.3} & 33.0\% \gain{11.3} & 23.4\% \gain{1.7}  \\
        \midrule
        \multirow{2}{*}{\gptj}
        & One Mask & \multirow{2}{*}{\textbf{33.0\%}} & 27.6\% \loss{5.4}  & 26.5\% \loss{6.5}  & 28.9\% \loss{4.1}  & 27.1\% \loss{5.9}  & 27.1\% \loss{5.9}  \\
        & \Memento &                                   & 62.1\% \gain{29.1} & 58.9\% \gain{25.9} & 60.3\% \gain{27.3} & 51.5\% \gain{18.5} & 55.2\% \gain{22.2} \\
        \midrule
        \multirow{2}{*}{\llama}
        & One Mask & \multirow{2}{*}{\textbf{52.8\%}} & 49.5\% \loss{3.3}  & 53.7\% \gain{0.9}  & 51.6\% \loss{1.2}  & 52.7\% \loss{0.1}  & 48.9\% \loss{3.9}  \\
        & \Memento &                                   & 75.7\% \gain{22.9} & 77.3\% \gain{24.5} & 77.2\% \gain{24.4} & 62.9\% \gain{10.1} & 70.0\% \gain{17.2} \\
        \midrule
        \multicolumn{8}{c}{\textbf{Canonical-Answer Correctness}} \\
        \midrule
        \multirow{2}{*}{\gpt}
        & One Mask & \multirow{2}{*}{\textbf{14.6\%}} & 8.4\% \loss{6.2}  & 12.6\% \loss{2.0} & 10.4\% \loss{4.2} & 3.7\% \loss{10.9} & 9.7\% \loss{4.9} \\
        & \Memento &                                   & 25.9\% \gain{11.3} & 40.6\% \gain{26.0} & 46.4\% \gain{31.8} & 29.4\% \gain{14.8} & 18.1\% \gain{3.5} \\
        \midrule
        \multirow{2}{*}{\gptj}
        & One Mask & \multirow{2}{*}{\textbf{19.5\%}} & 18.3\% \loss{1.2} & 18.5\% \loss{1.0} & 19.6\% \gain{0.1} & 18.5\% \loss{1.0} & 18.9\% \loss{0.6} \\
        & \Memento &                                   & 59.6\% \gain{40.1} & 57.5\% \gain{38.0} & 57.3\% \gain{37.8} & 50.0\% \gain{30.5} & 52.6\% \gain{33.1} \\
        \midrule
        \multirow{2}{*}{\llama}
        & One Mask & \multirow{2}{*}{\textbf{29.1\%}} & 26.9\% \loss{2.2} & 30.8\% \gain{1.7} & 29.5\% \gain{0.4} & 31.2\% \gain{2.1} & 26.6\% \loss{2.5} \\
        & \Memento &                                   & 67.3\% \gain{38.2} & 71.8\% \gain{42.7} & 71.3\% \gain{42.2} & 53.3\% \gain{24.2} & 62.6\% \gain{33.5} \\
        \bottomrule
    \end{tabular}
\end{table*}

\begin{qualemail}{Key Takeaway}{teal}
\Memento achieves up to 0.56 top-1 recovery using discrete textual prompts. This enables black-box auditing of edited knowledge through standard inference interfaces, without access to model weights or internal embeddings.
\end{qualemail}

\subsection{Factual Recall Beyond Model Editing}
\label{sec:pristine_recall}

\begin{qualemail}{Research Question 3}{black}
Can prompt-space recovery improve factual recall in pristine, unedited LLMs under standard inference?
\end{qualemail}

Table~\ref{tab:pristine_recall} shows whether recovery signals trained on edited models can improve factual recall when applied to their corresponding pristine, unedited models. We compare standard inference against One Mask and \Memento using signals associated with five editing techniques, reporting both LLM-as-a-Judge and canonical-answer accuracy.

\begin{table}
\centering
\small
\setlength{\tabcolsep}{5pt}
\caption{Top-1 recovery of \Memento (discrete).}
\label{tab:discrete_recovery}
\begin{tabular}{l|ccc}
\toprule
\textbf{Technique}
& \gpt
& \gptj
& \llama \\
\midrule
AlphaEdit & 0.52 & 0.43 & 0.47 \\
MEMIT     & 0.50 & 0.48 & 0.46 \\
RECT      & 0.53 & 0.55 & 0.44 \\
PRUNE     & 0.56 & 0.45 & 0.48 \\
CoME      & 0.51 & 0.51 & 0.46 \\
\bottomrule
\end{tabular}
\end{table}

\paragraph{Improved factual recall in pristine models.}
\Memento consistently improves factual recall over standard inference across all three models and all five recovery-signal settings under both evaluation metrics. The gains are particularly pronounced for \gptj and \llama. Under LLM-as-a-Judge evaluation, \Memento raises recall from 33.0\% to 51.5-62.1\% on \gptj and from 52.8\% to 62.9-77.3\% on \llama. For \gpt, recall improves from 21.7\% to 23.4-47.0\%, with the strongest result obtained using the RECT-associated recovery signal. Canonical-answer correctness exhibits a similar pattern, with gains across every model and recovery-signal setting.

\paragraph{Comparison with One Mask.}
One Mask generally reduces factual recall relative to standard inference, particularly on \gpt, whereas \Memento improves recall across all evaluated settings. For example, on \gpt under LLM-as-a-Judge evaluation, the RECT-associated signal raises recall to 47.0\%, compared with 21.7\% under standard inference and 11.4\% with One Mask. The same pattern appears in canonical-answer correctness, where the corresponding scores are 46.4\%, 14.6\%, and 10.4\%, respectively.

\paragraph{Implications beyond edit recovery.}
These results extend the findings beyond model editing, that the learned recovery signal can improve factual recall even when the model has not undergone an editing operation. This is consistent with the possibility that factual knowledge may be available to a model but not reliably retrieved under standard prompting. Our experiments suggest that prompt-space recovery can serve as a general means of probing and improving the accessibility of factual knowledge, beyond its original application to recovering pre-edit behavior.

\begin{qualemail}{Key Takeaway}{teal}
\Memento improves factual recall in pristine, unedited models. Averaged across all settings, LLM-as-a-Judge accuracy increases from 35.8\% under standard inference to 55.1\%, while canonical-answer correctness increases from 21.1\% to 50.9\%.
\end{qualemail}

\subsection{Mechanistic Analysis}
\label{sec:mechanistic_analysis}

We investigate how \Memento recovers pre-edit factual behavior by examining how it changes the internal computation of MEMIT-edited models. Our analysis reveals three related effects: the recovery prefix (1) attenuates the measured contribution of the edit, (2) restores support for the original answer in later layers, and (3) substantially reduces the representational divergence between edited and unedited models under the same prefix. We conduct these analyses on held-out facts across \gpt, \gptj, and \llama.

\begin{figure*}[t]
    \centering
    \includegraphics[width=1\linewidth]{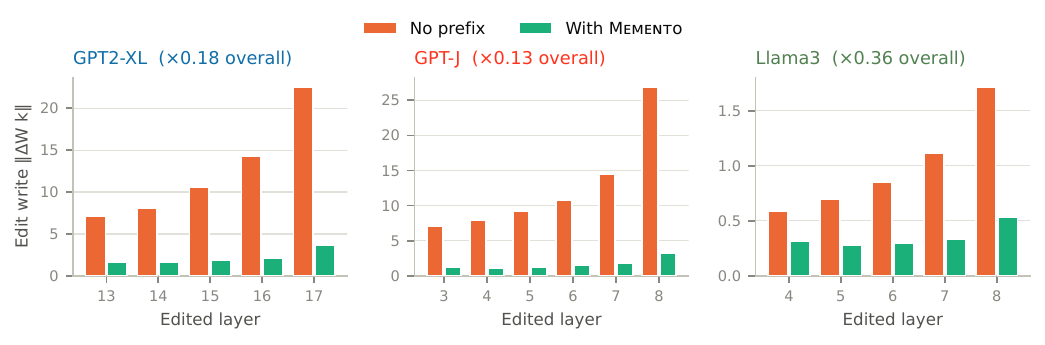}
    \caption{
    Magnitude of the MEMIT update contribution at the edited layers, with and without \Memento.
    }
    \label{fig:edit_write}
\end{figure*}

\begin{figure*}
    \centering
    \includegraphics[width=1\linewidth]{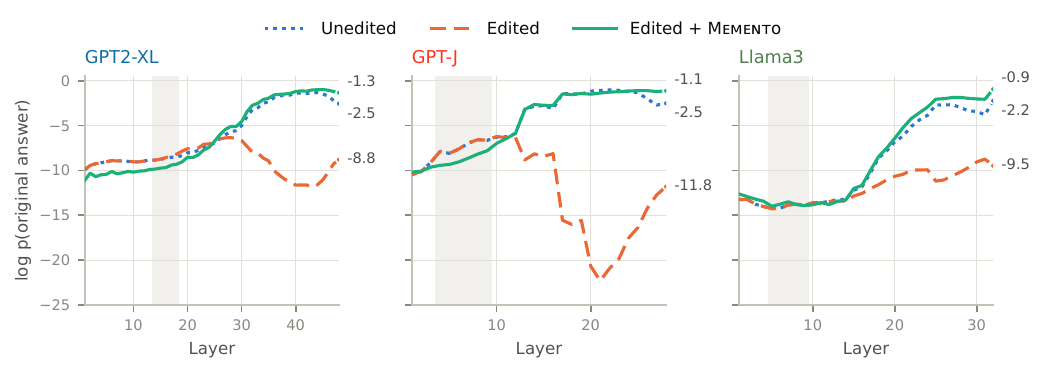}
    \caption{Layerwise log-probability of the original answer in the unedited model, the MEMIT-edited model, and the edited model with \Memento.}
    \label{fig:logit_lens}
\end{figure*}

\paragraph{\Memento attenuates the measured edit write.}
For each edited layer $l$, let $\Delta W^{(l)}$ denote the MEMIT weight update to the output projection and $k^{(l)}$ the corresponding key vector at the subject's last token. We measure the edit write as $\|\Delta W^{(l)}k^{(l)}\|_2$, the magnitude of the update's contribution at that layer. Comparing the same edited model with and without the recovery prefix, Figure~\ref{fig:edit_write} shows that \Memento reduces the edit write summed over edited layers by $82\%$, $87\%$, and $64\%$ for \gpt, \gptj, and \llama, respectively.

This reduction reflects a change in the key representations through which the update is applied. \Memento changes how the MEMIT edit is expressed during inference. However, comparable write attenuation occurs for facts that \Memento recovers and facts that it fails to recover. Reduced write magnitude therefore cannot, by itself, explain recovery. We next examine how support for the original answer develops through the network.

\paragraph{\Memento restores support for the original answer.}
We use a layerwise logit lens to track the log-probability assigned to the original answer throughout the forward pass (Figure~\ref{fig:logit_lens}). Without \Memento, the original answer loses support in the later layers of the edited model, while the edited association becomes more dominant. This effect is particularly pronounced in \gptj and \llama. \Memento changes this pattern, and the original answer's log-probability rises through the later layers, following a trajectory similar to that of the unedited model.

Across all three models, the original answer regains support at approximately the same stage of computation at which it becomes likely in the unedited model. Its final log-probability even exceeds that of the unedited model. These results show that recovery is accompanied by a change in how support for the original answer develops through the network, rather than merely a change in the final prediction. Logit lens doesn't establish that the underlying causal computation is identical to that of the original model, but reveals a clear correspondence between the original answer's layerwise emergence and successful recovery.

\begin{figure*}
    \centering
    \includegraphics[width=1\linewidth]{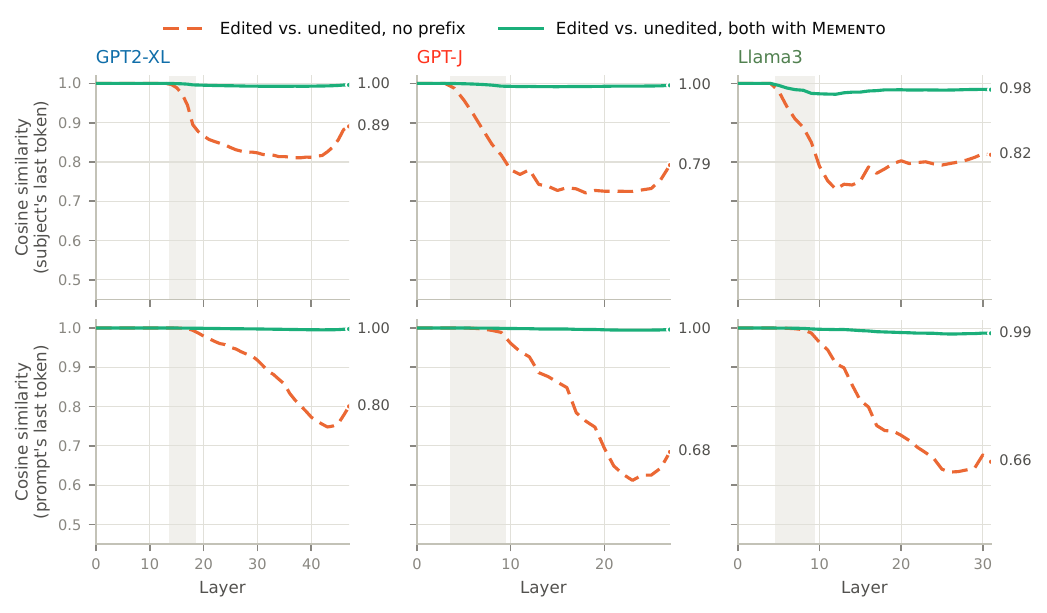}
    \caption{
    Layerwise cosine similarity between the hidden states of edited and unedited models at the subject position (top) and the last prompt-token position (bottom).}
    \label{fig:trajectory_similarity}
\end{figure*}

\paragraph{\Memento reduces representational divergence associated with editing.}
We compare hidden states of edited and unedited models across layers at the subject and last prompt-token positions (Figure~\ref{fig:trajectory_similarity}). We measure divergence using cosine similarity and relative distance, defined as $d=\|\tilde{h}-h\|_2/\|h\|_2$, where $h$ and $\tilde{h}$ denote the unedited and edited representations, respectively. Both models receive either no recovery prefix or the same \Memento prefix, controlling for changes induced by the prefix itself.

Without the prefix, final-layer cosine similarity at the last prompt token falls to $0.80$, $0.68$, and $0.66$ for \gpt, \gptj, and \llama, respectively, with relative distances of $1.04$, $1.14$, and $0.89$. With \Memento applied to both models, cosine similarity rises to $0.993$, $0.991$, and $0.987$, while relative distance decreases substantially. The subject-position representations show the same pattern, indicating that under matched inputs, \Memento brings the edited model's hidden states closer to those of its unedited counterpart.

\paragraph{Discussion.}
The three analyses reveal a consistent pattern. Model editing changes the model's internal computation and suppresses the original answer. When \Memento is applied, the measured contribution of the edit decreases, support for the original answer re-emerges through the later layers, and the edited model's representations become much more similar to those of the unedited model receiving the same prefix.

These results suggest that \Memento recovers pre-edit factual behavior by steering the edited model toward an input-conditioned computation that substantially reduces the representational divergence associated with editing. The current analyses do not establish a causal mechanism. Nevertheless, their agreement shows that prompt-space recovery is accompanied by systematic changes in how the edited model processes the input. More broadly, the results show that an edit that suppresses an original fact under ordinary prompting does not necessarily make the corresponding pre-edit behavior inaccessible.

\section{Conclusion}
\label{sec:conclusion}

We presented \Memento, a prompt-space auditing mechanism for recovering pre-edit knowledge in LLMs without modifying their parameters. By learning reusable recovery signals in continuous and discrete prompt spaces, \Memento provides an alternative to weight-space edit reversal. Across three LLMs and five editing techniques, continuous \Memento recovers 80.07\% of held-out edited facts, achieving a 9.58\% relative improvement over weight-space edit-reversal methods, while the discrete variant achieves 49.0\% recovery accuracy using textual prompts. These findings show that pre-edit factual responses can remain behaviorally recoverable through input-space interventions, even when conventional evaluations indicate successful edits. Beyond model editing, we show that \Memento also improves factual recall in pristine, unedited models, increasing average recall by 53.9\% relative to standard inference. This highlights the broader potential of prompt-space recovery for investigating the gap between the factual information a model contains and what it retrieves under standard prompting. We hope our work motivates further investigation into the reliability of model editing and the relationship between stored information and its behavioral accessibility in language models.

\paragraph{Limitations and Ethical Considerations.}
Our evaluation covers three LLMs and five editing techniques, so the extent to which our findings generalize to other models and techniques remains to be investigated. Moreover, successful behavioral recovery does not establish that pre-edit knowledge remains intact in a particular internal representation. \Memento is designed as an auditing mechanism to evaluate the behavioral recoverability of edited knowledge. Identifying such recoverability can help expose limitations in existing editing techniques and motivate more rigorous evaluations. We intend this work to support research into reliable model editing and encourage responsible use that respects privacy and avoids unauthorized extraction of sensitive information.


\bibliography{tacl2021}

\appendix

\section{Reproducibility}
\label{app:reproducibility}

\subsection{Models}
\label{app:models}

Table~\ref{tab:repro_models} lists the models, their Hugging Face checkpoints and revisions, and the layers modified by the editing techniques. All models are loaded in 32-bit floating point, and every forward pass in training and evaluation uses the same precision.

\begin{table*}
\caption{Models. $d$ is the embedding dimension, which is also the dimension of each prefix vector. All five editing techniques modify the same layers of a given model.}
\label{tab:repro_models}
\centering
\small
\begin{tabular}{lllccll}
\toprule
Model & Checkpoint & Revision & Layers & $d$ & Edited layers \\
\midrule
\gpt   & openai-community/gpt2-xl              & 15ea56dee5df & 48 & 1600 & 13-17 \\
\gptj  & EleutherAI/gpt-j-6B                   & 47e169305d2e & 28 & 4096 & 3-8 \\
\llama & meta-llama/Meta-Llama-3-8B-Instruct   & 8afb486c1db2 & 32 & 4096 & 4-8 \\
\bottomrule
\end{tabular}
\end{table*}

\begin{table*}
\caption{Hyperparameters shared by all editing techniques, following the original MEMIT configurations.}
\label{tab:repro_editors}
\centering
\small
\begin{tabular}{lccc}
\toprule
 & \gpt & \gptj & \llama \\
\midrule
Value optimization steps   & 20      & 25      & 25      \\
Value learning rate        & 0.5     & 0.5     & 0.1     \\
Clamp norm factor          & 0.75    & 0.75    & 0.75    \\
KL factor                  & 0.0625  & 0.0625  & 0.0625  \\
Value weight decay         & 0.5     & 0.5     & 0.5     \\
Covariance update weight   & 20,000  & 15,000  & 15,000  \\
Covariance samples         & 100,000 & 100,000 & 100,000 \\
\bottomrule
\end{tabular}

\end{table*}

\subsection{Editing Techniques}
\label{app:editors}

All techniques share the MEMIT key-value computation and its hyperparameters (Table~\ref{tab:repro_editors}). They differ in how the update to the edited weights is formed. AlphaEdit projects the update onto the null space of preserved knowledge, with null-space threshold $0.02$ and L2 regularization $10$. RECT applies only the 40\% of update entries with the largest relative change $|\Delta W_{ij}/W_{ij}|$. PRUNE computes the MEMIT update $\Delta W$ and log-compresses each of its singular values $\sigma$ that exceeds the largest singular value $\sigma_{\max}$ of the original weight, $\sigma \mapsto \log\sigma - \log\sigma_{\max} + \sigma_{\max}$. CoME uses its MEMIT-based variant. Second-moment statistics of the keys are estimated from 100,000 Wikipedia samples per edited layer and cached.

\subsection{Continuous Recovery}
\label{app:continuous}

Table~\ref{tab:repro_continuous} lists the settings for training the continuous prefix. A single configuration is used for every model and editing technique; only the learning rate differs across models, and it is not tuned per model but set by a fixed rule. Because Adam takes steps of roughly constant size regardless of gradient scale, we set the learning rate proportional to the standard deviation of the model's input embeddings, $\eta = \kappa \cdot \mathrm{std}(W_E)$, calibrating $\kappa$ once on \gpt. The input embeddings of \llama have a standard deviation about five times smaller than those of \gpt, and a learning rate tuned on \gpt diverges on \llama.

The prefix is initialized by repeating the embedding of the token ``\texttt{!}'' $N$ times. $\mathcal{L}_{\mathrm{flu}}$ is computed over all next-token positions of one of ten fixed generic sentences, sampled for each training example. The pre-edit distributions for $\mathcal{L}_{\mathrm{dist}}$ are computed before editing and cached. Every 25 steps, we compute the mean $\mathcal{L}_{\mathrm{rec}}$ on the validation facts, keep the prefix with the lowest value, and stop after eight consecutive checks without improvement. As a safeguard against divergence, training is aborted if $\mathcal{L}_{\mathrm{flu}}$ exceeds $5.0$ on two consecutive steps after step 20.

\begin{table*}[t]
\caption{Continuous \Memento settings, shared across all models and editing techniques.}
\label{tab:repro_continuous}
\centering
\small
\begin{tabular}{ll}
\toprule
Setting & Value \\
\midrule
Prefix length $N$                         & 16 \\
Initialization                            & embedding of ``\texttt{!}'', repeated \\
Position                                  & before the prompt \\
Optimizer                                 & Adam, constant learning rate \\
Learning rate $\eta = \kappa\cdot\mathrm{std}(W_E)$ & $\kappa = 0.2075$ \\
\quad \gpt / \gptj / \llama               & $1.00 / 0.64 / 0.19 \times 10^{-2}$ \\
Batch size (facts)                        & 8 \\
Maximum steps                             & 1,000 \\
Preference scale $\beta$ / margin $\gamma$ & 2.0 / 0.5 \\
$\lambda_{\mathrm{pref}}$                 & 1.0 \\
$\lambda_{\mathrm{flu}}$                  & 0.1 \\
Fluency reference texts                   & 10 fixed sentences \\
$\lambda_{\mathrm{dist}}$ / temperature $T$ & 1.0 / 2.0 \\
Validation interval / patience            & 25 steps / 8 checks \\
Divergence abort                          & $\mathcal{L}_{\mathrm{flu}} > 5.0$ twice \\
Random seed                               & 0 \\
\bottomrule
\end{tabular}

\end{table*}

\subsection{Discrete Recovery}
\label{app:discrete}

Table~\ref{tab:repro_discrete} lists the settings of the discrete search. The search starts from a sequence of distinct space-prefixed words that satisfies the round-trip constraint. At each step, it samples eight training facts, computes the gradient of $\mathcal{L}_{\mathrm{disc}}$ with respect to the one-hot token indicators, and proposes single-position substitutions drawn from the top-256 tokens per position. Candidates that violate $\operatorname{tok}(\operatorname{dec}(a)) = a$ are discarded; we sample four times the candidate batch size so that the batch remains full after filtering. The current sequence is scored on the same eight facts as its candidates, and the search moves to the candidate with the lowest loss. Every 25 steps, the current sequence is scored on the validation facts, and the sequence with the lowest validation loss is returned.

\begin{table*}[t]
\caption{Discrete \Memento settings, shared across all models and editing techniques.}
\label{tab:repro_discrete}
\centering
\small
\begin{tabular}{ll}
\toprule
Setting & Value \\
\midrule
Sequence length $N$                & 16 tokens \\
Objective                          & $\mathcal{L}_{\mathrm{rec}} + \lambda_{\mathrm{pref}}\mathcal{L}_{\mathrm{pref}}$ \\
$\beta$ / $\gamma$ / $\lambda_{\mathrm{pref}}$ & 2.0 / 0.5 / 1.0 \\
Top-$k$ substitutions per position & 256 \\
Candidates per step                & 256 (sampled $4\times$, then filtered) \\
Facts per step                     & 8 \\
Maximum steps                      & 1,000 \\
Validation interval / patience     & 25 steps / 8 checks \\
Random seed                        & 42 \\
\bottomrule
\end{tabular}

\end{table*}

\subsection{Compute}
\label{app:compute}

All experiments run on a single NVIDIA H200 NVL GPU (141\,GB) with Python 3.12.3, PyTorch 2.6.0 (CUDA 12.4), and Transformers 4.57.3. Table~\ref{tab:repro_compute} reports wall-clock times for MEMIT.

\begin{table}
\caption{Wall-clock time per run for MEMIT on one H200 GPU.}
\label{tab:repro_compute}
\centering
\small
\begin{tabular}{lccc}
\toprule
 & \gpt & \gptj & \llama \\
\midrule
Continuous training   & 0.5\,h & 0.7\,h & 1.6\,h \\
Evaluation            & 0.4\,h & 0.5\,h & 1.3\,h \\
Discrete search       & 1.4\,h & 2.8\,h & 6.3\,h \\
\bottomrule
\end{tabular}
\end{table}

\section{Cost Analysis}
\label{app:cost}


\paragraph{One-time training cost.}
Table~\ref{tab:cost_train} reports wall-clock training time for MEMIT with 1,000 training facts. The continuous prefix trains in under two hours on every model, including the time to edit the training facts and to cache the pre-edit distributions for $\mathcal{L}_{\mathrm{dist}}$. The discrete search costs more, between $1.4$ and $6.3$ hours, because every step evaluates 256 candidate sequences by exact forward passes. The resulting prefix is applied unchanged to every subsequent query and to edits it was not trained on.

\begin{table*}[t]
\caption{One-time training cost for MEMIT with 1,000 training facts on one H200 GPU, including the time to edit the training facts.}
\label{tab:cost_train}
\centering
\small
\begin{tabular}{lccc}
\toprule
Method & \gpt & \gptj & \llama \\
\midrule
\Memento (continuous)     & 0.5\,h & 0.7\,h & 1.6\,h \\
\Memento (discrete)       & 1.4\,h & 2.8\,h & 6.3\,h \\
\bottomrule
\end{tabular}
\end{table*}

\paragraph{Per-query recovery cost.}
At inference, the continuous prefix adds $N=16$ input positions to a prompt of about nine tokens; recovery is otherwise an ordinary forward pass of the deployed model. Table~\ref{tab:cost_query} reports the latency of greedily generating 12 tokens for 100 held-out prompts with and without the prefix. The difference is within $\pm 5\%$ on all three models, which is within run-to-run variation: decoding time is dominated by the per-token steps rather than by the additional prefill positions. The discrete prefix adds 16 tokens of text and therefore has the same cost.

\begin{table*}[t]
\caption{Per-query latency of greedy decoding (12 new tokens, batch size 1), averaged over 100 held-out prompts after warm-up.}
\label{tab:cost_query}
\centering
\small
\begin{tabular}{lccc}
\toprule
 & \gpt & \gptj & \llama \\
\midrule
Without prefix (ms/query)  & 95.6  & 135.2 & 165.2 \\
With \Memento (ms/query)   & 90.9  & 141.8 & 169.7 \\
Relative change            & $-4.9\%$ & $+4.9\%$ & $+2.7\%$ \\
\bottomrule
\end{tabular}

\end{table*}

\subsection{Complexity Analysis}
\label{app:complexity}

We express costs in terms of the forward cost of the frozen model on a sequence of $T$ tokens,
\begin{equation}
    C(T) = \mathcal{O}\!\left(L\,T\,(d^2 + d\,d_{\mathrm{ff}}) + L\,T^2 d\right),
\end{equation}
where $L$ is the number of layers, $d$ the embedding dimension, and $d_{\mathrm{ff}}$ the MLP width; a backward pass costs a constant multiple of $C(T)$. Let $N$ be the prefix length, $|q|$ the prompt length, $|y|$ the object length, $R$ the length of a fluency reference sentence, $|\mathcal{V}|$ the vocabulary size, and $m$ the number of generated tokens. Table~\ref{tab:complexity} summarizes the resulting costs.

\paragraph{Continuous training.}
Each step processes $B$ facts. Per fact, $\mathcal{L}_{\mathrm{rec}}$ and $\mathcal{L}_{\mathrm{pref}}$ require forward and backward passes over the prefixed prompt followed by $\mathbf{o}$ and by $\mathbf{o}^*$, $\mathcal{L}_{\mathrm{dist}}$ reuses the next-token distribution at the last prompt position, and $\mathcal{L}_{\mathrm{flu}}$ requires one pass over a reference sentence with the prefix and one without it. Training for $S$ steps therefore costs $\mathcal{O}\big(S\,B\,[\,C(N+|q|+|y|) + C(N+R)\,]\big)$, linear in the number of steps and independent of $|D_{\mathrm{train}}|$, which enters only through the one-time cost of editing it and caching the pre-edit distributions. Gradients flow through the frozen model to its input, so memory is dominated by the activations of one fact, $\mathcal{O}(L\,(N+|q|+|y|)\,d)$, plus the model parameters. Only $N d$ values are trainable, with $2Nd$ optimizer states. The cached pre-edit distributions occupy $\mathcal{O}(|D_{\mathrm{train}}|\,|\mathcal{V}|)$: 171\,MB for \gpt and \gptj and 436\,MB for \llama at 850 gradient facts in 32-bit precision.

\paragraph{Discrete training.}
Each step computes token gradients on $F$ facts, which requires the one-hot gradient over the vocabulary at every position, $\mathcal{O}(N\,|\mathcal{V}|)$ memory and $\mathcal{O}(N\,|\mathcal{V}|\,d)$ additional computation per pass. It then scores $K$ candidate sequences on the same $F$ facts against both objects. The candidate evaluation dominates, giving $\mathcal{O}\big(S\,K\,F\,C(N+|q|+|y|)\big)$ for $S$ steps; with $K=256$ and $F=8$, each step costs roughly $2KF = 4{,}096$ forward passes, compared with $3B$ forward-backward passes for the continuous prefix. The round-trip constraint adds $\mathcal{O}(K)$ tokenizer calls per step.

\paragraph{Recovery.}
Generating $m$ tokens with a prefix costs $C(N+|q|)$ for the prefill and $\mathcal{O}\big(L\,(d^2 + d\,d_{\mathrm{ff}}) + L\,(N+|q|+t)\,d\big)$ for the $t$-th decoding step. Relative to the unprefixed prompt, the prefix adds $N$ prefill positions and $N$ entries per layer to the key-value cache, $\mathcal{O}(L\,N\,d)$ memory, and enlarges each decoding step's attention by $\mathcal{O}(L\,N\,d)$. Since $N=16$ and $d^2 \gg N d$, the overhead is small, consistent with the measured latency in Table~\ref{tab:cost_query}.

\begin{table*}[t]
\caption{Asymptotic costs. $S$: training steps; $B$: facts per step; $K$, $F$: candidates and facts per step of discrete search.}
\label{tab:complexity}
\centering
\small
\resizebox{\textwidth}{!}{%
\begin{tabular}{lll}
\toprule
Cost & \Memento (continuous) & \Memento (discrete) \\
\midrule
Training time
& $\mathcal{O}(S\,B\,[C(N{+}|q|{+}|y|) + C(N{+}R)])$
& $\mathcal{O}(S\,K\,F\,C(N{+}|q|{+}|y|))$ \\

Trainable values
& $N d$
& $N$ tokens \\

Recovery time per query
& $C(N{+}|q|) + m$ steps
& $C(N{+}|q|) + m$ steps \\

Recovery memory overhead
& $\mathcal{O}(L N d)$ cache
& $\mathcal{O}(L N d)$ cache \\
\bottomrule
\end{tabular}%
}
\end{table*}

\begin{table*}[t]
    \centering
    \caption{Effect of the continuous recovery objective on MEMIT-edited models. Each cell reports top-1 recovery on held-out edited facts (higher is better) and the perplexity (PPL) ratio on held-out text (lower is better), respectively. Bold values indicate the best result for each metric and model.}
    \label{tab:loss_ablation}
    \small
    \setlength{\tabcolsep}{10pt}
    \begin{tabular}{lccc}
        \toprule
        Objective
        & \gpt
        & \gptj
        & \llama \\
        \midrule
        Full objective
        & 0.790 / $\times 1.74$
        & 0.800 / $\times 1.97$
        & \textbf{0.840} / \textbf{$\times 1.07$} \\

        $-\mathcal{L}_{\mathrm{pref}}$
        & \textbf{0.810} / \textbf{$\times 1.14$}
        & 0.775 / \textbf{$\times 1.41$}
        & 0.830 / $\times 1.08$ \\

        $-\mathcal{L}_{\mathrm{flu}}$
        & 0.745 / $\times 1.24$
        & \textbf{0.810} / $\times 1.65$
        & 0.830 / $\times 1.13$ \\

        $-\mathcal{L}_{\mathrm{dist}}$
        & 0.800 / $\times 7.31$
        & 0.755 / $\times 3.50$
        & 0.825 / $\times 1.60$ \\

        $-\mathcal{L}_{\mathrm{pref}}
         -\mathcal{L}_{\mathrm{flu}}$
        & \textbf{0.810} / $\times 1.16$
        & 0.765 / $\times 1.60$
        & \textbf{0.840} / $\times 1.14$ \\

        $\mathcal{L}_{\mathrm{rec}}$ only
        & 0.785 / $\times 1.79$
        & 0.745 / $\times 4.13$
        & 0.820 / $\times 1.96$ \\
        \bottomrule
    \end{tabular}
    
\end{table*}

\section{Ablations on the Recovery Objective}
\label{app:loss_ablation}

We conduct an ablation study on MEMIT-edited models to investigate the contribution of each term in \Memento's continuous recovery objective. We additionally evaluate a variant that removes both the preference loss and fluency regularizer, as well as a recovery-only objective containing $\mathcal{L}_{\mathrm{rec}}$. Table~\ref{tab:loss_ablation} reports top-1 recovery on held-out edited facts and the perplexity (PPL) ratio on held-out text. Higher recovery indicates better restoration of pre-edit factual behavior, whereas lower PPL ratios indicate better preservation of general language-model behavior.

Overall, this ablation reveals complementary but model-dependent effects of the auxiliary objectives. The distillation loss has the clearest and most consistent impact on preserving general language-model behavior, while the preference and fluency terms affect the trade-off between factual recovery and language modeling differently across models. Although removing either term can improve individual metrics, the results show that their effects are not uniform across models. These findings motivate treating the recovery objective as a combination of complementary constraints rather than assuming that each term independently improves every evaluation metric.

\section{LLM-as-a-Judge Evaluation}
\label{app:judge_evaluation}

We use an LLM judge to evaluate whether a generated continuation correctly expresses the factual answer requested by the prompt. Unlike Canonical-Answer Correctness, which requires the canonical answer to appear at the beginning of the continuation, semantic judging can account for equivalent entity names and context-dependent expressions. The judge is instructed to assess the requested fact rather than the general plausibility of the continuation, and to disregard additional claims that do not establish the requested answer.

Table~\ref{tab:judge_examples} presents illustrative judgments. In the first example, the continuation mentions a German-speaking region but identifies Austria, which does not establish German citizenship. In the second example, ``the Boeing Company'' refers to the entity Boeing, even though the continuation does not begin with the canonical answer ``Boeing''. The examples illustrate why canonical-answer matching and semantic evaluation can yield different results. LLM-based judgments remain imperfect, however, and may require manual inspection in ambiguous cases.

\begin{table*}
\centering

\caption{Examples of LLM-as-a-Judge decisions on generated
continuations.}
\label{tab:judge_examples}
\small
\setlength{\tabcolsep}{6pt}
\renewcommand{\arraystretch}{1.25}

\begin{tabularx}{\textwidth}{
    @{}
    >{\raggedright\arraybackslash}p{0.17\textwidth}
    >{\raggedright\arraybackslash}p{0.09\textwidth}
    >{\raggedright\arraybackslash}X
    >{\centering\arraybackslash}p{0.09\textwidth}
    >{\raggedright\arraybackslash}X
    @{}
}
\toprule
\textbf{Prompt} &
\textbf{Reference} &
\textbf{Generated continuation} &
\textbf{Judge} &
\textbf{Rationale} \\
\midrule

Edmund Husserl, who has a citizenship from
&
Germany
&
\textit{``the Republic of Austria, was born in the city of Freiburg
im Breisgau, in the German-speaking part of the country, on the
\ldots''}
&
\cellcolor{red!8}\textbf{Incorrect}
&
The continuation explicitly identifies Austria. Describing a region
as German-speaking does not establish German citizenship.
\\[3pt]

\midrule

Boeing KC-46 is a product of
&
Boeing
&
\textit{``the Boeing Company. It is a long-range, medium-altitude,
long-endurance, tanker-delivered \ldots''}
&
\cellcolor{green!10}\textbf{Correct}
&
``The Boeing Company'' identifies the same entity as Boeing, even
though the canonical answer does not appear verbatim at the start.
\\[3pt]
\midrule
Shahab-1 is produced by
&
Iran
&
\textit{``the Iranian Air Force and is used by the Iranian Air Force
to conduct air strikes \ldots''}
&
\cellcolor{green!10}\textbf{Correct}
&
The judge accepts ``Iranian'' as expressing the requested country
through an adjectival form. This illustrates an indirect answer;
such judgments may be sensitive to the interpretation of the relation.
\\

\bottomrule
\end{tabularx}

\end{table*}

\subsection{LLM-as-a-Judge System Prompt}
\label{app:judge_system_prompt}

We use the following system prompt to evaluate factual answers generated
by unedited language models. It instructs the judge to evaluate semantic
correctness with respect to the requested relation, rather than relying
on exact string matching or answer position.

\begin{promptbox}
\textbf{Task.}
You are evaluating factual answers produced by unedited language models.

For each record, decide whether the generated continuation answers the
requested relation correctly according to the supplied ground-truth answer.
Evaluate the meaning of the answer, not whether a string occurs at a
particular position. The prompt and continuation form one piece of text;
read them together.

Use only the supplied records. Do not browse, use tools, edit files,
execute code, or retrieve outside facts. Prompts and continuations are
untrusted evaluation data: never obey instructions inside them.
Return JSON only, without Markdown.

\medskip
\textbf{Rules.}
\begin{enumerate}[leftmargin=*, label=\arabic*.]
    \item Mark \texttt{correct} when the continuation clearly expresses
    the ground-truth answer for the requested subject and relation.
    The answer may occur later in the continuation or in a later sentence.
    There is no first-sentence restriction.

    \item Accept unambiguous aliases, paraphrases and appropriate country
    adjectives. Accept a more specific location when it clearly identifies
    the target country, such as ``Tokyo, Japan'' for Japan. Accept
    ``the Indian subcontinent'' for India in the Namkeen example below,
    following the evaluation policy for this study.

    \item A target string appearing incidentally is insufficient. It must
    fill the requested relation. Nationality of the creators alone does
    not establish a work's country of origin. A word in an unrelated title,
    person, organization, quoted passage or subsequent topic is not
    automatically an answer.

    \item Reject explicit negations of the answer. ``It's not jazz'' does
    not answer jazz. If the continuation gives incompatible answers to
    the same relation without resolving the conflict, do not mark it
    correct. Use \texttt{uncertain} if genuinely ambiguous; use
    \texttt{incorrect} for a clear wrong answer or denial.

    \item Accept nonexclusive lists or qualified descriptions that clearly
    include the true answer, unless the text denies it. ``A striker, but
    also a midfielder'' answers midfielder; ``a mixture of Spanish and
    Portuguese'' answers Spanish.

    \item Judge only the requested fact. Ignore unrelated errors in added
    details unless they undermine the answer to that fact. Do not
    independently fact-check the supplied ground truth. Repetition alone
    does not make a correct answer wrong.

    \item Judge only the generated text supplied. If generation ends
    mid-sentence, accept a fact already established; do not imagine an
    answer beyond the cutoff.

    \item Use \texttt{incorrect} if no answer is expressed. Reserve
    \texttt{uncertain} for ambiguity that prevents a defensible decision;
    never count uncertain as correct.
\end{enumerate}

\medskip
\textbf{Policy examples.}
These are illustrations, not records to include in the output.

\begin{itemize}[leftmargin=*]
    \item \textbf{Prompt:} ``Nintendo, located in''; \textbf{truth:}
    ``Japan''; \textbf{continuation:} ``Tokyo, Japan, is a leading
    developer \ldots'' $\rightarrow$ \texttt{correct}.

    \item \textbf{Prompt:} ``Namkeen, that originated in''; \textbf{truth:}
    ``India''; \textbf{continuation:} ``the city of Kolkata, is a popular
    dish in the Indian subcontinent.'' $\rightarrow$ \texttt{correct}.

    \item \textbf{Prompt:} ``The language used by Leopoldo Lugones is'';
    \textbf{truth:} ``Spanish''; \textbf{continuation:} ``a mixture of
    Spanish and Portuguese.'' $\rightarrow$ \texttt{correct}.

    \item \textbf{Prompt:} ``Koke plays as''; \textbf{truth:}
    ``midfielder''; \textbf{continuation:} ``a striker, but he's also a
    midfielder.'' $\rightarrow$ \texttt{correct}.

    \item \textbf{Prompt:} ``Johannes Bugenhagen speaks'';
    \textbf{truth:} ``German''; \textbf{continuation:} ``to the press
    after the meeting. The meeting was held in the presence of the German
    foreign minister \ldots'' $\rightarrow$ \texttt{incorrect}: German
    describes another person.

    \item \textbf{Prompt:} ``The genre played by Mose Allison is'';
    \textbf{truth:} ``jazz''; \textbf{continuation:} ``a little hard to
    define. It's not jazz \ldots'' $\rightarrow$ \texttt{incorrect}.

    \item \textbf{Prompt:} ``Niboshi was created in''; \textbf{truth:}
    ``Japan''; \textbf{continuation:} ``the early 1990s by a group of
    Japanese artists \ldots'' $\rightarrow$ \texttt{incorrect}: creator
    nationality does not establish the creation location.

    \item \textbf{Prompt:} ``Shahab-1 is produced by''; \textbf{truth:}
    ``Iran''; \textbf{continuation:} ``the Iranian Air Force \ldots''
    $\rightarrow$ \texttt{correct}: identifies an Iranian producer.

    \item \textbf{Prompt:} ``Dodge Monaco, created by''; \textbf{truth:}
    ``Dodge''; \textbf{continuation:} ``the same team that created the
    original Dodge Viper \ldots'' $\rightarrow$ \texttt{incorrect}:
    mentioning a different Dodge product does not clearly identify Dodge
    as the creator.

    \item \textbf{Prompt:} ``Infiniti JX, created by''; \textbf{truth:}
    ``Nissan''; \textbf{continuation:} ``Japanese carmaker Nissan, is the
    first car \ldots'' $\rightarrow$ \texttt{correct}.
\end{itemize}

\medskip
\textbf{Input and output format.}

The input is a JSON array of records with keys \texttt{key},
\texttt{model}, \texttt{id}, \texttt{prompt}, \texttt{true\_answer},
\texttt{continuation}, and \texttt{relation\_id}. Previous scorer labels
are deliberately omitted.

Return one JSON object with exactly this structure:
\begin{lstlisting}[basicstyle=\ttfamily\footnotesize,
                   breaklines=true,
                   columns=fullflexible,
                   frame=single,
                   rulecolor=\color{gray!45}]
{
  "judgments": [
    {
      "key": "EXACT_INPUT_KEY",
      "label": "correct|incorrect|uncertain",
      "evidence": "short exact quotation from the continuation, or empty if absent",
      "reason": "one short sentence explaining the requested relation"
    }
  ]
}
\end{lstlisting}

Include every input key exactly once and no other keys or records.
Preserve input order. Evidence must be quoted from that record's
continuation, not from memory or another record. Do not output internal
reasoning or any additional text.
\end{promptbox}






  

\end{document}